\documentclass[twoside,11pt]{article}

\usepackage{jmlr2e_mod}
\usepackage{amsmath}
\usepackage{natbib}
\usepackage[bottom]{footmisc}

\jmlrheading{1}{2017}{1-48}{4/00}{10/00}{Jared Ostmeyer and Lindsay Cowell}

\ShortHeadings{Machine Learning on Sequential Data Using a Recurrent Weighted Average}{Ostmeyer and Cowell}
\firstpageno{1}

\begin{document}

\title{Machine Learning on Sequential Data Using a Recurrent Weighted Average}

\author{\name Jared Ostmeyer * \email jostmey@gmail.com \\
       \name Lindsay Cowell \email Lindsay.Cowell@UTSouthwestern.edu \\
       \addr Department of Clinical Sciences\\
       UT Southwestern Medical Center\\
       5323 Harry Hines Blvd.\\
       Dallas, TX 75390-9066, USA \\ \\
       \textnormal{* Indicates primary correspondence}}


\maketitle

\begin{abstract}
Recurrent Neural Networks (RNN) are a type of statistical model designed to handle sequential data. The model reads a sequence one symbol at a time. Each symbol is processed based on information collected from the previous symbols. With existing RNN architectures, each symbol is processed using only information from the previous processing step. To overcome this limitation, we propose a new kind of RNN model that computes a recurrent weighted average (RWA) over every past processing step. Because the RWA can be computed as a running average, the computational overhead scales like that of any other RNN architecture. The approach essentially reformulates a simplified form of the \textit{attention mechanism}, one that has no query, into a stand-alone model. The performance of the RWA model is assessed on the variable copy problem, the adding problem, classification of artificial grammar, classification of sequences by length, and classification of the MNIST images (\textit{where the pixels are read sequentially one at a time}). On almost every task, the RWA model is found to outperform a standard LSTM model.
\end{abstract}

\begin{center}
  Source code and experiments at https://github.com/jostmey/rwa
\end{center}


\section{Introduction}

\begin{figure}[!b]
  \center
  \includegraphics[width=13.75cm,height=5.0cm]{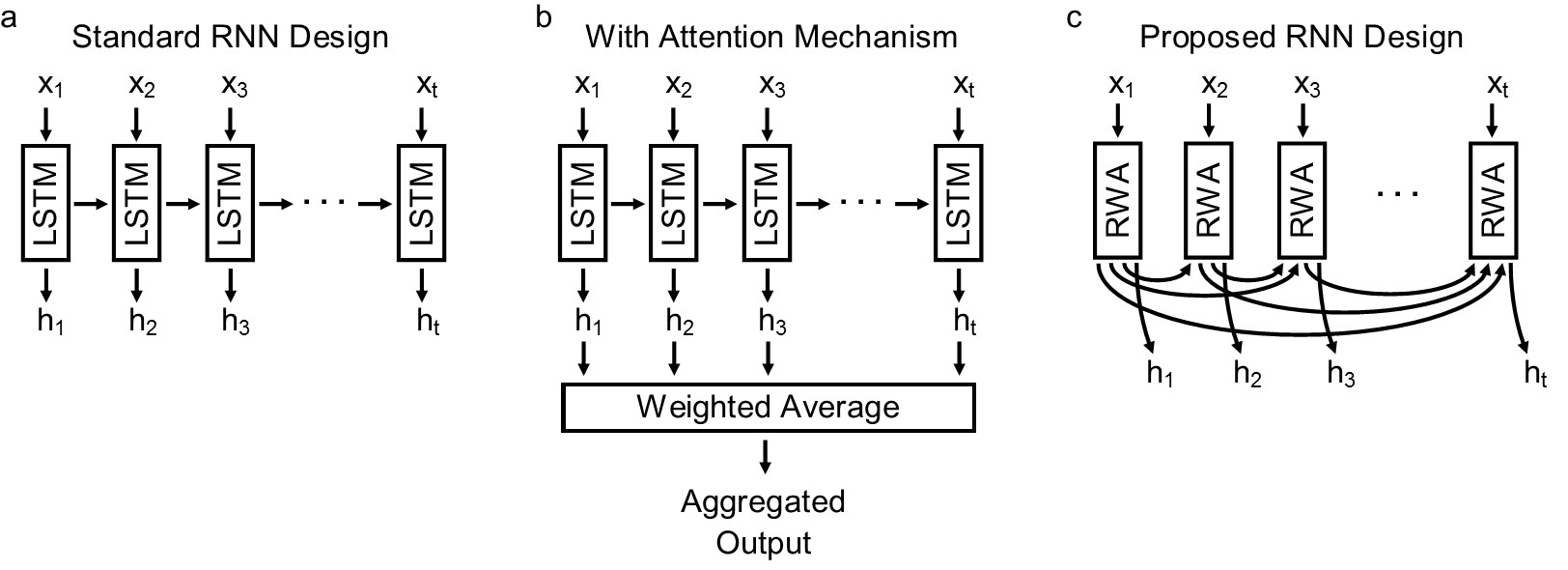}
  \caption[Comparison of RNN architectures]{\label{fig:intro}Comparison of models for classifying sequential data. (a) Standard RNN architecture with LSTM requires that information contained in the first symbol $x_1$ pass through the feedback connections repeatedly to reach the output $h_t$, like in a game of \textit{telephone} (\textit{a.k.a. Chinese whispers}). (b) A simplified form of the attention mechanism aggregates the outputs into a single state by computing a weighted average. It is not recursively defined. (c) The proposed model incorporates pathways to every previous processing step using a recurrent weighted average (RWA). By maintaining a running average, the computational cost scales like that of other RNN models.}
\end{figure}

Types of information as dissimilar as language, music, and genomes can be represented as sequential data. The essential property of sequential data is that the order of the information is important, which is why statistical algorithms designed to handle this kind of data must be able to process each symbol in the order that it appears. Recurrent neural network (RNN) models have been gaining interest as a statistical tool for dealing with the complexities of sequential data. The essential property of a RNN is the use of feedback connections. The sequence is read by the RNN one symbol at a time through the model's inputs. The RNN starts by reading the first symbol and processing the information it contains. The processed information is then passed through a set of feedback connections. Every subsequent symbol read into the model is processed based on the information conveyed through the feedback connections. Each time another symbol is read, the processed information of that symbol is used to update the information conveyed in the feedback connections. The process continues until every symbol has been read into the model (Fig. \ref{fig:intro}a). The processed information is passed along each step like in the game \textit{telephone} (\textit{a.k.a. Chinese whispers}). With each step, the RNN produces an output that serves as the model's prediction. The challenge of designing a working RNN is to make sure that processed information does not decay over the many steps. Error correcting information must also be able to \textit{backpropagate} through the same pathways without degradation \citep*{1,2}. Hochreiter and Schmidhuber were the first to solve these issues by equipping a RNN with what they called long short-term memory (LSTM) \citep*{3}.

Since the introduction of the LSTM model, several improvements have been proposed. The \textit{attention mechanism} is perhaps one of the most significant \citep*{4}. The attention mechanism is essentially a weighted average. At each step, the output from the RNN is weighted by an \textit{attention model}, creating a weighted output. The weighted outputs are then aggregated together by computing a weighted average (Fig. \ref{fig:intro}b). The outcome of the weighted average is used as the model's result. The attention model controls the relative contribution of each output, determining how much of each output is ``seen" in the results. Usually, the attention model is also given a \textit{query} that tells it what to look for, and the weights are recalculated each time the query changes \citep*{4}. A simplified form of the attention mechanism has no query, so each weight depends only on the output being weighted \citep*{7,20,21,22}. The attention mechanism has since been incorporated into several other neural network architectures leading to a variety of new models each specifically designed for a single task (partial reference list: \citealt{4,6,7,8,9}). Unfortunately, the attention mechanism is not defined in a recurrent manner. The recurrent connections must come from a separate RNN model, restricting where the attention mechanism can be used.

Inspired by the attention mechanism used to modify existing neural network architectures, we propose a new kind of RNN that is created by reformulating the simplified form of the attention mechanism into a stand-alone model. The proposed model includes feedback connections from every past processing step, not just the preceding step (Fig. \ref{fig:intro}c). The feedback connections from each processing step are weighted by an attention model. The weighted feedback connections are then aggregated together by computing a weighted average. The model is said to use a recurrent weighted average (RWA) because the attention model also uses the feedback connections. At each step, the weighted average can be computed as a running average. This is possible because there is no query, so each weight is fixed when its symbol is read. By maintaining a running average, the model scales like any other type of RNN.

\section{The RWA Model}

\subsection{Mathematical Description}

The proposed model is defined recursively over the length of the sequence. Starting at the first processing step $t=1$ a set of values $h_0$, representing the initial state of the model, is required. The distribution of values for $h_0$ must be carefully chosen to ensure that the initial state resembles the output from the subsequent processing steps. This is accomplished by defining $h_0$ in terms of $s_0$, a parameter that must be fitted to the data.

\begin{equation}
 \label{eq:1}
 h_0=f(s_0)
\end{equation}

\noindent The parameters $s_0$ are passed through the model's activation function $f$ to mimic the processes that generate the outputs of the later processing steps.

For every processing step that follows, a weighted average computed over every previous step is passed through the activation function $f$ to generate a new output $h_t$ for the model. The equation for the model is given below.

\begin{equation}
 \label{eq:2}
 h_t = f\Bigg(\frac{\sum\limits_{i=1}^t z(x_i, h_{i-1}) \circ e^{a(x_i, h_{i-1})}}{\sum\limits_{j=1}^t e^{a(x_j, h_{j-1})}}\Bigg)
\end{equation}

\noindent The weighted average consists of two models: $z$ and $a$. The model $z$ encodes the features $x_i$ for each symbol in the sequence. Its recurrence relations, represented by $h_{i-1}$, provide the context necessary to encode the features in a sequence dependent manner. The model $a$ serves as the attention model, determining the relative contribution of $z$ at each processing step. The model $a$ is not given a query. It depends only on $x_i$ and $h_{i-1}$, so the weight $e^{a(x_i, h_{i-1})}$ for step $i$ is fixed when $x_i$ is read. The exponential terms of model $a$ are normalized in the denominator to form a proper weighted average. The recurrent relations in $a$, represented by $h_{i-1}$, are required to compose the weighted average recursively. Because of the recurrent terms in $a$, the model is said to use a RWA.

There are several models worth considering for $z$, but only one is considered here. Because $z$ encodes the features, the output from $z$ should ideally be dominated by the values in $x_i$ and not $h_{i-1}$. This can be accomplished by separating the model for $z$ into an unbounded component containing only $x_i$ and a bounded component that includes the recurrent terms $h_{i-1}$.

\begin{equation}
 \label{eq:3}
 z(x_i, h_{i-1}) = u(x_i) \circ \tanh{g(x_i, h_{i-1})}
\end{equation}

\noindent The model for $u$ contains only $x_i$ and encodes the features. With each processing step, information from $u$ can accumulate in the RWA. The model for $g$ contains the recurrent relations and is bounded between $[-1,1]$ by the $\tanh$  function. This model can control the sign of $z$ but cannot cause the absolute magnitude of $z$ to increase. Having a separate model for controlling the sign of $z$ ensures that information encoded by $u$ does not just accumulate but can negate information encoded from previous processing steps. Together the models $u$ and $g$ encode the features in a sequence dependent manner.

The terms $u$, $g$, and $a$ can be modelled as feed-forward linear networks.

\begin{equation}
 \label{eq:4}
 \begin{split}
   &u(x_i) = W_u \cdot x_i+b_u \\
   &g(x_i,h_{i-1}) = W_g \cdot [x_i,h_{i-1}]+b_g \\
   &a(x_i,h_{i-1}) = W_a \cdot [x_i,h_{i-1}]
 \end{split}
\end{equation}

\noindent The matrices $W_u$, $W_g$, and $W_a$ represent the weights of the feed-forward networks, and the vectors $b_u$ and $b_g$ represent the bias terms. The bias term for $a$ would cancel when factored out of the numerator and denominator, which is why the term is omitted.

While running through a sequence, the output $h_t$ from each processing step can be passed through a fully connected neural network layer to predict a label. Gradient descent based methods can then be used to fit the model parameters, minimizing the error between the true and predicted label.

\subsection{Running Average}

The RWA in equation (\ref{eq:2}) is recalculated from the beginning at each processing step. The first step to reformulate the model as a running average is to separate the RWA in equation (\ref{eq:2}) as a numerator term $n_t$ and denominator term $d_t$.

\begin{equation*}
 \begin{split}
   &n_t = \sum\limits_{i=1}^t z(x_i, h_{i-1}) \circ e^{a(x_i, h_{i-1})} \\
   &d_t = \sum\limits_{j=1}^t e^{a(x_j, h_{j-1})}
 \end{split}
\end{equation*}

\noindent Because any summation can be rewritten as a recurrence relation, the summations for $n_t$ and $d_t$ can be defined recurrently (see Appendix A). Let $n_0=0$ and $d_0=0$.

\begin{equation}
 \label{eq:5}
 \begin{split}
   &n_t = n_{t-1}+z(x_t, h_{t-1}) \circ e^{a(x_t, h_{t-1})} \\
   &d_t = d_{t-1}+e^{a(x_t, h_{t-1})}
 \end{split}
\end{equation}

\noindent By saving the previous state of the numerator $n_{t-1}$ and denominator $d_{t-1}$, the values for $n_t$ and $d_t$ can be efficiently computed using the work done during the previous processing step. The output $h_t$ from equation (\ref{eq:2}) can now be obtained from the relationship listed below.

\begin{equation}
 \label{eq:6}
 h_t = f\bigg(\frac{n_t}{d_t}\bigg)
\end{equation}

\noindent Using this formulation of the model, the RWA can efficiently be computed dynamically.

\subsection{Equations for Implementation}

The RWA model can be implemented using equations (\ref{eq:1}) and (\ref{eq:3})--(\ref{eq:6}), which are collected together and written below.

\begin{equation}
 \label{eq:7}
 \begin{split}
   &h_0 = f(s_0) \ , \ \ n_0 = 0 \ , \ \  d_0 = 0 \\
   &u(x_t) = W_u \cdot x_t+b_u \\
   &g(x_t,h_{t-1}) = W_g \cdot [x_t,h_{t-1}]+b_g \\
   &a(x_t,h_{t-1}) = W_a \cdot [x_t,h_{t-1}] \\
   &z(x_t, h_{t-1}) = u(x_t) \circ \tanh{g(x_t, h_{t-1})} \\
   &n_t = n_{t-1}+z(x_t, h_{t-1}) \circ e^{a(x_t, h_{t-1})} \\
   &d_t = d_{t-1}+e^{a(x_t, h_{t-1})} \\
   &h_t = f\bigg(\frac{n_t}{d_t}\bigg)
 \end{split}
\end{equation}

\noindent Starting from the initial conditions, the model is run recursively over an entire sequence. The features for every symbol in the sequence are contained in $x_t$, and the parameters $s_0$, $W_u$, $b_u$, $W_g$, $b_g$, and $W_a$ are determined by fitting the model to a set of training data. Because the model is differentiable, the parameters can be fitted using gradient optimization techniques.

In most cases, the numerator and denominator will need to be rescaled to prevent the exponential terms from becoming too large or small. The rescaling equations are provided in Appendix B.

\section{Experiments}

\subsection{Implementations of the Models}

A RWA model is implemented in \textit{TensorFlow} using the equations in (\ref{eq:7}) \citep*{10}. The model is trained and tested on five different classification tasks each described separately in the following subsections.

The same configuration of the RWA model is used on each dataset. The activation function is $f(x)=\tanh{x}$ and the model contains $250$ units. Following general guidelines for initializing the parameters of any neural network, the initial weights in $W_u$, $W_g$ and $W_a$ are drawn at random from the uniform distribution $\Big[-\sqrt{\frac{3}{(N_\text{in}+N_\text{out})/2}},\sqrt{\frac{3}{(N_\text{in}+N_\text{out})/2}}\Big]$ and the bias terms $b_u$ and $b_g$ are initialized to $0$'s \citep*{11}. The initial state $s_0$ for the RWA model is drawn at random from a normal distribution according to $s_0 \sim \mathcal{N}(\mu=0,\sigma^2=1)$. To avoid not-a-number (NaN) and divide-by-zero errors, the numerator and denominator terms in equations (\ref{eq:7}) are rescaled using equations (\ref{eq:9}) in Appendix B, which do not alter the model's output.

The datasets are also scored on a LSTM model that contains $250$ cells to match the number of units in the RWA model. Following the same guidelines used for the RWA model, the initial values for all the weights are drawn at random from the uniform distribution $\Big[-\sqrt{\frac{3}{(N_\text{in}+N_\text{out})/2}},\sqrt{\frac{3}{(N_\text{in}+N_\text{out})/2}}\Big]$ and the bias terms are initialized to 0's except for the forget gates \citep*{11}. The bias terms of the forget gates are initialized to $1$'s, since this has been shown to enhance the performance of LSTM models \citep*{12, 13}. All initial cell states of the LSTM model are $0$.

A fully connected neural network layer transforms the output from the $250$ units into a predicted label. The error between the true label and predicted label is then minimized by fitting the model's parameters using Adam optimization \citep*{14}. All values for the ADAM optimizer follow published recommended settings. A step size of $0.001$ is used throughout this study, and the other optimizer settings are $\beta_1=0.9$, $\beta_2=0.999$, and $\epsilon=10^{-8}$. Each parameter update consists of a batch of 100 training examples. Gradient clipping is not used.

Each model is immediately scored on the test set and no negative results are omitted\footnote{The RWA model was initially implemented incorrectly. The mistake was discovered by Alex Nichol. After correcting the mistake, the old results were discarded and the RWA model was run again on each classification task.}. No hyperparameter search is done and no regularization is tried. At every $100$ steps of training, a batch of $100$ test samples are scored by each model to generate values that are plotted in the figures for each of the tasks described below. The code and results for each experiment may be found online (\textit{see:} https://github.com/jostmey/rwa).

\subsection{Classifying Artificial Grammar}

\begin{figure}[!t]
  \center
  \includegraphics[width=11.5cm,height=3.75cm]{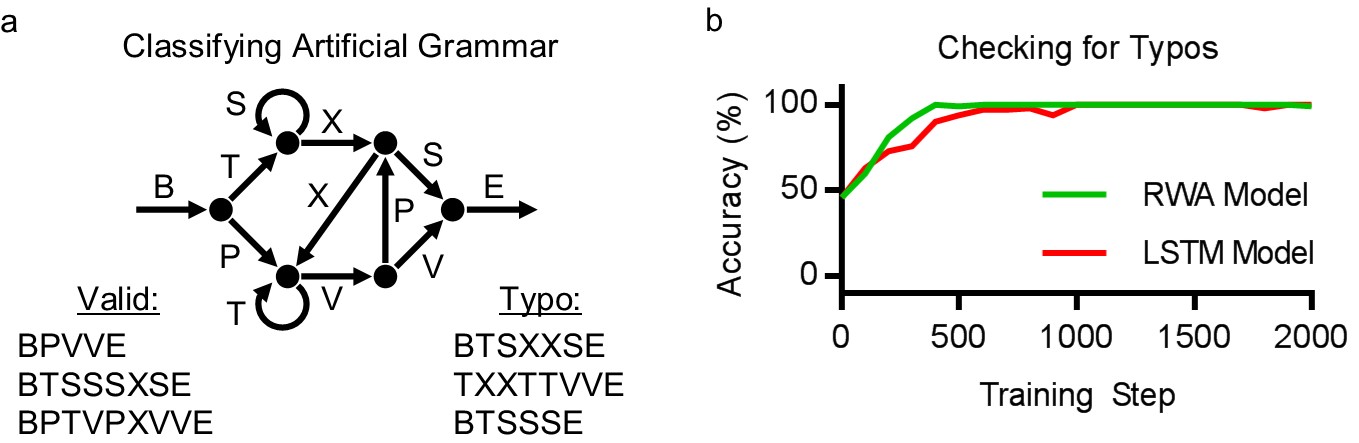}
  \caption[Classifying artificial grammar]{\label{fig:reber}(a) The generator function used to create each sentence. Examples of valid and invalid sentences are shown below. (b) A plot comparing the performance of the RWA and LSTM models. The traces show the accuracy of each model on the test data while the models are being fitted to the training data. The RWA model reaches $100$\% accuracy slightly before the LSTM model.}
\end{figure}

It is important that a model designed to process sequential data exhibit a sensitivity to the order of the symbols. For this reason, the RWA model is tasked with proofreading the syntax of sentences generated by an artificial grammatical system. Whenever the sentences are valid with respect to the artificial grammar, the model must return a value of 1, and whenever a typo exists the model must return a value of 0. This type of task is considered especially easy for standard RNN models, and is included here to show that the RWA model also performs well at this task \citep*{3}.

The artificial grammar generator is shown in Figure \ref{fig:reber}a. The process starts with the arrow labeled \textit{B}, which is always the first letter in the sentence. Whenever a node is encountered, the next arrow is chosen at random. Every time an arrow is used the associated letter is added to the sentence. The process continues until the last arrow \textit{E} is used. All valid sentences end with this letter. Invalid sentences are constructed by randomly inserting a typo along the sequence. A typo is created by an invalid jump between unconnected arrows. No more than one typo is inserted per sentence. Typos are inserted into approximately half the sentences. Each RNN model must perform with greater than $50$\% accuracy to demonstrate it has learned the task.

A training set of $100,000$ samples are used to fit the model, and a test set of $10,000$ samples are used to evaluate model performance. The RWA model does remarkably well, achieving $100$\% accuracy in $600$ training steps. The LSTM model also learns to identify valid sentences, but requires $1000$ training steps to achieve the same performance level (Fig \ref{fig:reber}b). This task demonstrates that the RWA model can classify patterns based on the order of information.

\subsection{Classifying by Sequence Length}

Classifying sequences by length requires that RNN models track the number of symbols contained in each sequence. For this task, the length of each sequence is randomly drawn from a uniform distribution over every possible length $0$ to $T$, where $T$ is the maximum possible length of the sequence. Each step in the sequence is populated with a random number drawn from a unit normal distribution (\textit{i.e. $\mu=0$ and $\sigma^2=1$}). Sequences greater than length $T/2$ are labeled with $1$ while shorter sequences are labeled with $0$. The goal is to predict these labels, which indicates if a RNN model has the capacity to classify sequences by length. Because approximately half the sequences will have a length above $T/2$, each RNN model must perform with greater than $50$\% accuracy to demonstrate it has learned the task.

\begin{figure}[!t]
  \center
  \includegraphics[width=5.5cm,height=3.5cm]{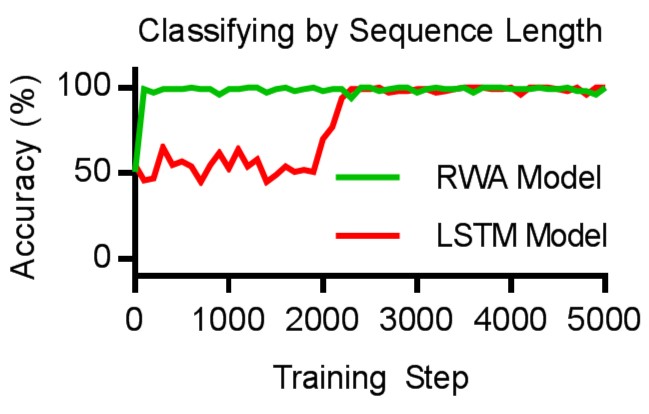}
  \caption[Classifying by sequence length]{\label{fig:counting_problem} Plot comparing the performance of the RWA and LSTM models when classifying sequences by length. The models must determine if a sequence is longer than $500$ symbols, which is half the maximum possible sequence length. Each sequence is populated with numbers drawn at random from a unit normal distribution. The traces show the accuracy of each model on the test data while the models are being fitted to the training data. The RWA model achieves a classification accuracy near $100$\% before the LSTM model.}
\end{figure}

For this task, $T=1,000$ (\textit{The task was found to be too easy for both models for $T=100$}). A training set of $100,000$ samples are used to fit the model, and a test set of $10,000$ samples are used to evaluate model performance. The RWA model does remarkably well, learning to correctly classify sequences by their length in fewer than 100 training steps. The LSTM model also learns to correctly classify sequences by their length, but requires over 2,000 training steps to achieve the same level of performance (Fig. \ref{fig:counting_problem}).

\subsection{Variable Copy Problem}

The variable copy problem, proposed by \citet{15}, requires that a RNN memorize a random sequence of symbols and recall it only when prompted. The input sequence starts with the recall sequence. The recall sequence for the RNN to memorize consists of $S$ many symbols drawn at random from $\{a_i\}_{i=1}^K$. After that, the input sequence contains a stretch of $T$ blank spaces, denoted by the symbol $a_{K+1}$. Somewhere along this stretch of blank spaces, one of the symbols is replaced by a delimiter $a_{K+2}$. The delimiter indicates when the recall sequence should be repeated back. The input sequence is then padded with an additional stretch of $S$ blank spaces, providing sufficient time for the model to repeat the recall sequence. The goal is to train the model so that its output is always blank except for the spaces immediately following the delimiter, in which case the output must match the recall sequence.

\begin{figure}[!t]
  \center
  \includegraphics[width=15cm,height=5.5cm]{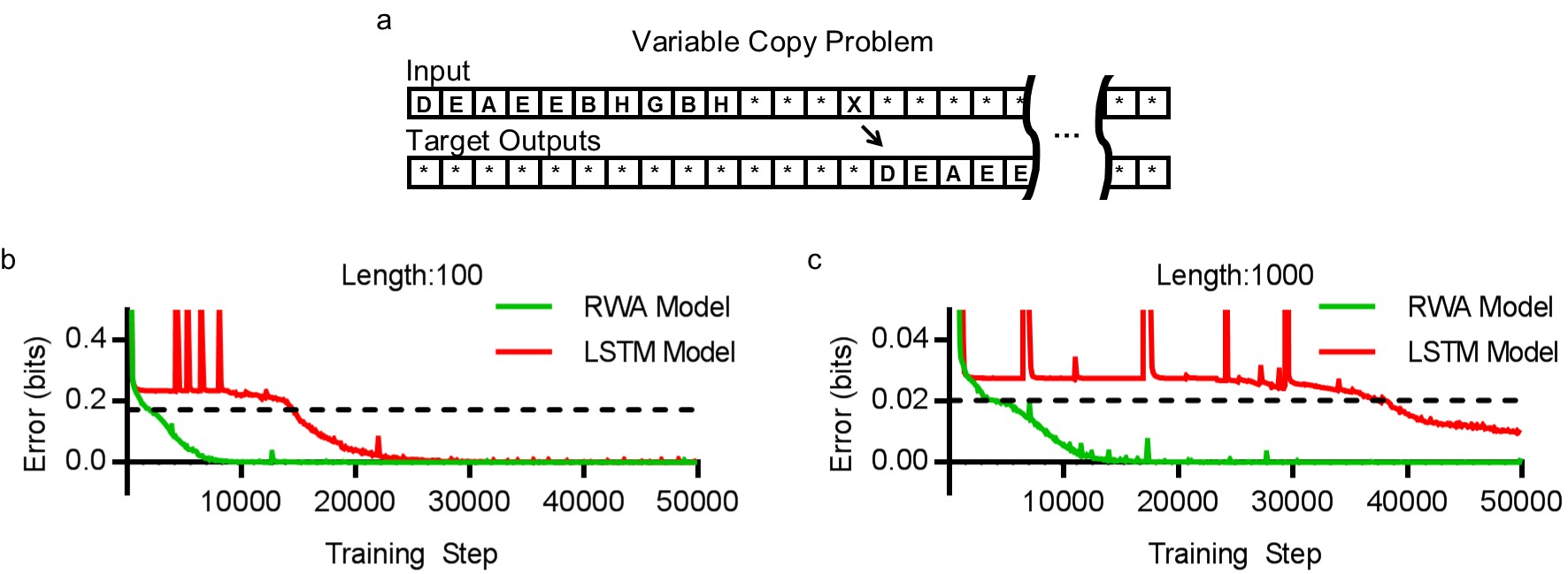}
  \caption[Variable copy problem]{\label{fig:copy_problem}(a) An example of the variable copy problem. The input and target sequences represent a single sample of data. (b) A plot comparing the performance of the RWA and LSTM models when the sequences are $100$ symbols in length. The traces show the error of each model on the test data while the models are being fitted to the training data. The RWA model beats the baseline score (dashed line) before the LSTM model. (c) The same problem as before except the sequences are $1,000$ symbols in length. The RWA model beats the baseline score (dashed line) before the LSTM model.}
\end{figure}

An example of the task is shown in Figure \ref{fig:copy_problem}a. The recall pattern drawn at random from symbols \textit{A} through \textit{H} is \textit{DEAEEBHGBH}. Blank spaces represented by \textit{*} fill the rest of the sequence. One blank space is chosen at random and replaced by \textit{X}, which denotes the delimiter. After \textit{X} appears the model must repeat the recall pattern in the output. The na\"{i}ve strategy is to always guess that the output is \textit{*} because this is the most common symbol. Each RNN must perform better than the na\"{i}ve strategy to demonstrate that it has learned the task. Using this na\"{i}ve strategy, the expected cross-entropy error between the true output and predicted output is $\frac{S \cdot \ln{k}}{2 \cdot S +T}$, represented by the dashed line in Figures \ref{fig:copy_problem}b, c. This is the baseline to beat.

For this challenge, $K=8$, $S=10$, and models are trained and evaluated on two separate cases where $T=100$ and $T=1,000$. For both $T=100$ and $T=1,000$, the training set contains $100,000$ examples and the test set contains $10,000$ examples. For the case of $T=100$, the RWA requires roughly $1,000$ training steps to beat the baseline score, whereas the LSTM model requires over $10,000$ training steps to achieve the same level of performance. The RWA model scales well to $T=1,000$, requiring only $3,000$ training steps to beat the baseline score. The LSTM model is only barely able to beat the baseline error after $50,000$ training steps (Fig. \ref{fig:copy_problem}c). The RWA model appears to scale much better as the sequence length increases.

\subsection{Adding Problem}

The adding problem, first proposed by \citet{3}, tests the ability of a RNN model to form long-range connections across a sequence. The task requires that the model learn to add two numbers randomly spaced apart on a sequence. The input sequence consists of two values at each step. The first value serves as an indicator marking the value to add while the second value is the actual number to be added and is drawn at random from a uniform distribution over $[0,1]$. Whenever the indicator has a value of $1$, the randomly drawn number must be added to the sum, and whenever the indicator has a value of $0$, the randomly drawn number must be ignored. Only two steps in the entire sequence will have an indicator of $1$, leaving the indicator $0$ everywhere else.

\begin{figure}[!t]
  \center
  \includegraphics[width=15cm,height=5.5cm]{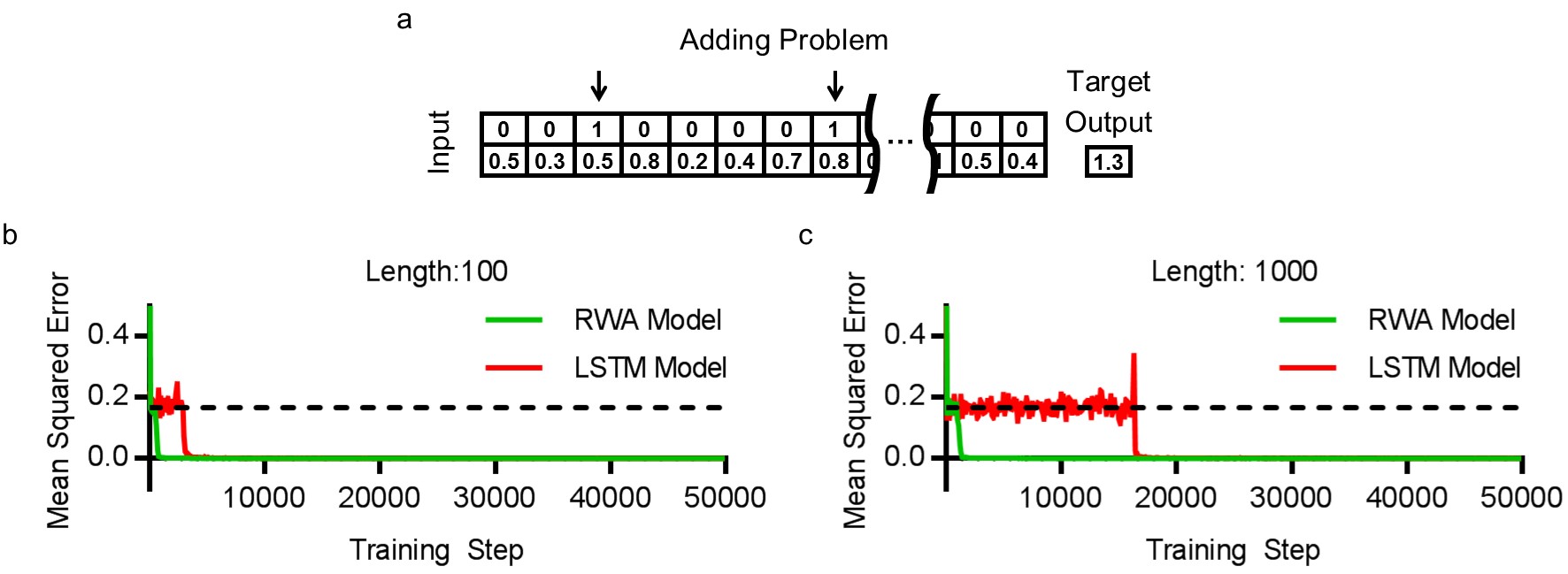}
  \caption[Adding problem]{\label{fig:adding_problem}(a) An example of the adding problem. The input sequence and target output represent a single sample of data. (b) A plot comparing the performance of the RWA and LSTM models when the sequences are of length $100$. The traces show the error of each model on the test data while the models are being fitted to the training data. The RWA model beats the baseline score (dashed line) before the LSTM model. (c) The same problem as before except the sequences are of length $1,000$. The RWA model beats the baseline score (dashed line) before the LSTM model.}
\end{figure}

An example of the adding problem is shown in Figure \ref{fig:adding_problem}a. The top row contains the indicator values and the bottom row contains randomly drawn numbers. The two numbers that have an indicator of 1 must be added together. The numbers in this example are $0.5$ and $0.8$, making the target output $1.3$. Because the two numbers being added together are uniformly sampled over $[0,1]$, the na\"{i}ve strategy is to always guess that the target output is $1$. Each RNN must perform better than the na\"{i}ve strategy to demonstrate that it has learned the task. Using this na\"{i}ve strategy, the expected mean square error (MSE) between the true answer and the prediction is approximately $0.167$, represented by the dashed line in Figures \ref{fig:adding_problem}b,c. This is the baseline to beat.

The adding problem is repeated twice, first with sequences of length $100$ and again with sequences of length $1,000$. In both cases, a training set of $100,000$ samples are used to fit the model, and a test set of $10,000$ samples are used to evaluate the model’s performance. When the sequences are of length $100$, the RWA model requires fewer than $1,000$ training steps to beat the baseline score while the LSTM model requires around $3,000$ steps (Fig. \ref{fig:adding_problem}b). When the sequences are of length $1,000$, the RWA model requires approximately $1,000$ training steps to beat the baseline score, while the LSTM model requires over $15,000$ training steps on the same task (Fig. \ref{fig:adding_problem}c). The RWA model appears to scale much better as the sequence length increases.

\subsection{Classifying MNIST Images (Pixel by Pixel)}

The MNIST dataset contains $28\times28$ pixel images of handwritten digits 0 through 9. The goal is to predict the digit represented in the image \citep*{16}. Using the same setup suggested by \citet*{17}, the images are arranged into a sequence of pixels. The length of each sequence is $28\times28=784$ pixels. Each RNN model reads the sequence one pixel at a time and must predict the digit being represented in the image from this sequence.

\begin{figure}[!t]
  \center
  \includegraphics[width=15cm,height=5.5cm]{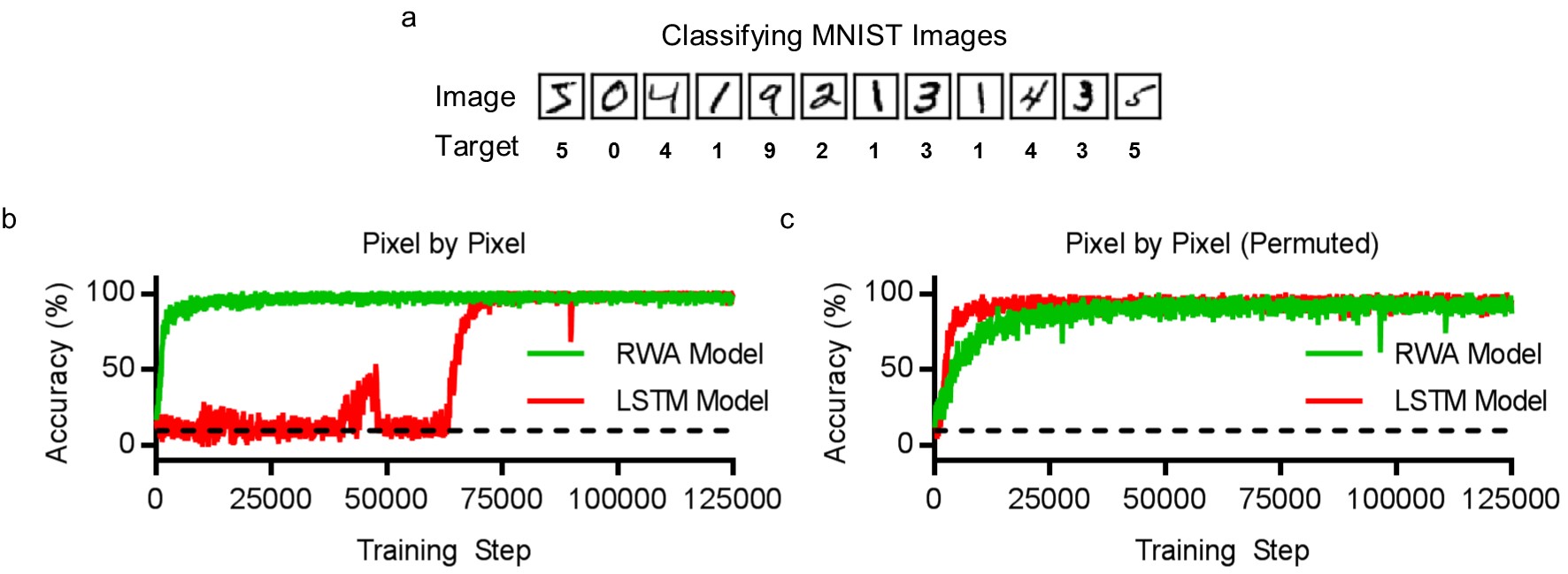}
  \caption[Classifying MNIST images]{\label{fig:mnist}(a) Examples of the MNIST classification task. Each image of a handwritten digit must be classified by the value it represents. The images are feed into the RNNs as a sequence of pixels one at a time. (b) A plot comparing the performance of the RWA and LSTM models. The traces show the accuracy of each model on the test data while the models are being fitted to the training data. The RWA model beats the baseline score (dashed line) before the LSTM model. (c) Same task as before expect that the pixels have been randomly permuted. The LSTM model trained much faster on the permutation task while the RWA model took slightly longer.}
\end{figure}

Examples of MNIST digits with the correct label are shown in Figure \ref{fig:mnist}a. The pixels at the top and bottom of each image are empty. When the images are arranged into a sequence of pixels, all the important pixels will be in the middle of the sequence. To utilize these pixels, each RNN model will need to form long-range dependencies that reach the middle of each sequence. The model will have formed the necessary long-range dependencies when it outperforms a na\"{i}ve strategy of randomly guessing each digit. A na\"{i}ve strategy will achieve an expected accuracy of $10$\%, represented by the dashed line in Figures \ref{fig:mnist}b. This is the baseline to beat.

For this challenge, the  standard training set of $60,000$ images is used to fit the model, and the standard test set of $10,000$ images is used to evaluate the model’s performance on unseen data. The RWA model fits the dataset in under $20,000$ steps, considerably faster than the LSTM model (Fig. \ref{fig:mnist}b). After a quarter million training steps, the RWA model achieves an accuracy of $98.1$\% with an error of $0.175$ bits, while LSTM model achieves an accuracy of $99.0$\% with an error of $0.077$ bits. In this example, the LSTM model generalizes better to the unseen data.

A separate and more challenging task is to randomly permute the pixels, leaving the pixels out of order, as described in \citet{17}. The same permutation mapping must be used on each image to keep the data consistent between images. As before, a na\"{i}ve strategy of randomly guessing the answer will achieve an expected accuracy of $10$\%, represented by the dashed line in Figures \ref{fig:mnist}c. This is the baseline to beat.

The classification task is repeated with the pixels randomly permuted. This time the LSTM model fit the dataset faster than the RWA model (Fig. \ref{fig:mnist}c). After a quarter million training steps, the RWA model achieves an accuracy of $93.5$\% with an error of $0.561$ bits, while LSTM model achieves an accuracy of $93.6$\% with an error of $0.577$ bits. Neither model generalizes noticeably better to the unseen data.

\section{Discussion}

The RWA model reformulates a simplified form of the attention mechanism, one that has no query, into a stand-alone model that can be optimized using gradient descent based methods. Given that the attention mechanism has been shown to work well on a wide range of problems, even in its simplified form, the robust performance of the RWA model on the five classification tasks in this study is not surprising \citep*{4,6,7,8,9,20,21,22}. Moreover, the RWA model did not require a hyperparameter search to tailor the model to each task. The same configuration successfully generalized to unseen data on every task. Clearly, the RWA model can form long-range dependencies across the sequences in each task and does not suffer from the vanishing or exploding gradient problem that affects other RNN models \citep*{1,2}.

The RWA model requires less clock time and fewer parameters than a LSTM model with the same number of units. On almost every task, the RWA model beat the baseline score using fewer training steps. The number of training steps could be further reduced using a larger step size for the parameter updates. It is unclear how large the step size can become before the convergence of the RWA model becomes unstable (\textit{a larger step size may or may not require gradient clipping, which was not used in this study}). The RWA model also uses over $25$\% fewer parameters per unit than a LSTM model. Depending on the computer hardware, the RWA model can either run faster or contain more units than a LSTM model on the same computer.

Unlike previous implementations of the attention mechanism that read an entire sequence before generating a result, the RWA model computes an output in real time with each new input. With this flexibility, the RWA model can be deployed anywhere existing RNN models can be used. Several architectures are worth exploring. Bidirectional RWA models for interpreting genomic data could be created to simultaneously account for information that is both upstream and downstream in a sequence. Multi-layered versions could also be created to handle XOR classification problems at each processing step. In addition, RWA models could be used to autoencode sequences or to learn mappings from a fixed set of features to a sequence of labels. The RWA model offers a compelling framework for performing machine learning on sequential data.

\section{Conclusion}

The RWA model opens exciting new areas for research. Because the RWA model can form direct pathways to any past processing step, it can detect patterns in a sequence that other models would miss. This could lead to dramatically different outcomes when applied to complex tasks like natural language processing, automated music composition, and the classification of genomic sequences. Given how easily the model can be inserted into existing RNN architectures, it is worth trying the RWA model on these tasks. The RWA model has the potential to solve problems that have been deemed too difficult until now.


\acks{Special thanks are owed to Elizabeth Harris for proofreading and editing the manuscript. She brought an element of clarity to the manuscript that it would otherwise lack. Alex Nichol also needs to be recognized. Alex identified a flaw in equations (\ref{eq:9}), which are used to compute a numerically stable update of the numerator and denominator terms. Without Alex's careful examination of the manuscript, all the results for the RWA model would be incorrect. The department of Clinical Sciences at the University of Texas Southwestern Medical Center also needs to be acknowledged. Ongoing research projects at the medical center highlighted the need to find better ways to process genomic data and provided the inspiration behind the development of the RWA model.

This work was supported by a training grant from the Cancer Prevention and Research Institute of Texas (RP160157) and an R01 from the National Institute of Allergy and Infectious Diseases (R01AI097403).}


\newpage

\appendix

\section*{Appendix A.}
\label{app:recursive}


Any summation of the form $y_n=\sum\limits_{i=1}^n r_i$ can be written as a recurrent relation. Let the initial values be $y_0=0$.

\begin{equation*}
 y_n = \sum\limits_{i=1}^n r_i = \sum\limits_{i=1}^{n-1} r_i + r_n = y_{n-1}+r_n
\end{equation*}

\noindent The summation is now defined recursively.

\section*{Appendix B.}
\label{app:rescale}


The exponential terms in equations (\ref{eq:7}) are prone to underflow and overflow. The underflow condition can cause all the terms in the denominator to become zero, leading to a divide-by-zero error. The overflow condition leads to not-a-number (NaN) errors. To avoid both kinds of errors, the numerator and denominator can be multiplied by an \textit{exponential scaling factor}. Because the numerator and denominator are scaled by the same factor, the quotient remains unchanged and the output of the model is identical. The idea is similar to how a softmax function must sometimes be rescaled to avoid the same errors. 

The exponential scaling factor is determined by the largest value in $a$ among every processing step. Let $a^\text{max}_t$ represent the largest observed value. The initial value for $a^\text{max}_0$ needs to be less than any value that will be observed, which can be accomplished by using an extremely large negative number.

\begin{equation}
 \label{eq:9}
 \begin{split}
  a&^\text{max}_0 = -10^{38} \\ 
  a&^\text{max}_t = \text{Max}\big\{a^\text{max}_{t-1}, a(x_t,h_{t-1})\big\} \\
  n&_t = n_{t-1} \circ e^{a^\text{max}_{t-1}-a^\text{max}_t} + z(x_t, h_{t-1}) \circ e^{a(x_t, h_{t-1})-a^\text{max}_t} \\
  d&_t = d_{t-1} \circ e^{a^\text{max}_{t-1}-a^\text{max}_t} + e^{a(x_t, h_{t-1})-a^\text{max}_t}
 \end{split}
\end{equation}

\noindent The first equation sets the initial value for the exponential scaling factor to one of the largest numbers that can be represented using single-precision floating-point numbers. Starting with this value avoids the underflow condition. The second equation saves the largest value observed in $a$. The final two equations compute an updated numerator and denominator. The equations scale the numerator and denominator accounting for the exponential scaling factor used during previous processing steps. The results from equations (\ref{eq:9}) can replace the results for $n_t$ and $d_t$ in equations (\ref{eq:7}) without affecting the model's output $h_t = f\big(\frac{n_t}{d_t}\big)$.

\vskip 0.2in
\bibliography{references}

\begin{thebibliography}{19}
\providecommand{\natexlab}[1]{#1}
\providecommand{\url}[1]{\texttt{#1}}
\expandafter\ifx\csname urlstyle\endcsname\relax
  \providecommand{\doi}[1]{doi: #1}\else
  \providecommand{\doi}{doi: \begingroup \urlstyle{rm}\Url}\fi

\bibitem[Abadi et~al.(2016)]{10}
Abadi et~al.
\newblock Tensorflow: Large-scale machine learning on heterogeneous distributed
  systems.
\newblock \emph{arXiv preprint arXiv:1603.04467}, 2016.

\bibitem[Bahdanau et~al.(2014)Bahdanau, Cho, and Bengio]{4}
Dzmitry Bahdanau, Kyunghyun Cho, and Yoshua Bengio.
\newblock Neural machine translation by jointly learning to align and
  translate.
\newblock \emph{arXiv preprint arXiv:1409.0473}, 2014.

\bibitem[Bengio et~al.(1994)Bengio, Simard, and Frasconi]{2}
Yoshua Bengio, Patrice Simard, and Paolo Frasconi.
\newblock Learning long-term dependencies with gradient descent is difficult.
\newblock \emph{IEEE transactions on neural networks}, 5\penalty0 (2):\penalty0
  157--166, 1994.

\bibitem[Chan et~al.(2015)Chan, Jaitly, Le, and Vinyals]{8}
William Chan, Navdeep Jaitly, Quoc~V Le, and Oriol Vinyals.
\newblock Listen, attend and spell.
\newblock \emph{arXiv preprint arXiv:1508.01211}, 2015.

\bibitem[Gers et~al.(2000)Gers, Schmidhuber, and Cummins]{12}
Felix~A Gers, J{\"u}rgen Schmidhuber, and Fred Cummins.
\newblock Learning to forget: Continual prediction with lstm.
\newblock \emph{Neural computation}, 12\penalty0 (10):\penalty0 2451--2471,
  2000.

\bibitem[Glorot and Bengio(2010)]{11}
Xavier Glorot and Yoshua Bengio.
\newblock Understanding the difficulty of training deep feedforward neural
  networks.
\newblock In \emph{Aistats}, volume~9, pages 249--256, 2010.

\bibitem[Henaff et~al.(2016)Henaff, Szlam, and LeCun]{15}
Mikael Henaff, Arthur Szlam, and Yann LeCun.
\newblock Recurrent orthogonal networks and long-memory tasks.
\newblock In \emph{Proceedings of The 33rd International Conference on Machine
  Learning}, pages 2034--2042, 2016.

\bibitem[Hochreiter(1991)]{1}
Sepp Hochreiter.
\newblock \emph{Untersuchungen zu dynamischen neuronalen Netzen}.
\newblock PhD thesis, diploma thesis, institut f{\"u}r informatik, lehrstuhl
  prof. brauer, technische universit{\"a}t m{\"u}nchen, 1991.

\bibitem[Hochreiter and Schmidhuber(1997)]{3}
Sepp Hochreiter and J{\"u}rgen Schmidhuber.
\newblock Long short-term memory.
\newblock \emph{Neural computation}, 9\penalty0 (8):\penalty0 1735--1780, 1997.

\bibitem[Jozefowicz et~al.(2015)Jozefowicz, Zaremba, and Sutskever]{13}
Rafal Jozefowicz, Wojciech Zaremba, and Ilya Sutskever.
\newblock An empirical exploration of recurrent network architectures.
\newblock In \emph{Proceedings of The 32nd International Conference on Machine
  Learning}, pages 2342--2350, 2015.

\bibitem[Kingma and Ba(2014)]{14}
Diederik Kingma and Jimmy Ba.
\newblock Adam: A method for stochastic optimization.
\newblock \emph{arXiv preprint arXiv:1412.6980}, 2014.

\bibitem[Le et~al.(2015)Le, Jaitly, and Hinton]{17}
Quoc~V Le, Navdeep Jaitly, and Geoffrey~E Hinton.
\newblock A simple way to initialize recurrent networks of rectified linear
  units.
\newblock \emph{arXiv preprint arXiv:1504.00941}, 2015.

\bibitem[LeCun et~al.(1998)LeCun, Bottou, Bengio, and Haffner]{16}
Yann LeCun, L{\'e}on Bottou, Yoshua Bengio, and Patrick Haffner.
\newblock Gradient-based learning applied to document recognition.
\newblock \emph{Proceedings of the IEEE}, 86\penalty0 (11):\penalty0
  2278--2324, 1998.

\bibitem[Lin et~al.(2017)Lin, Feng, dos Santos, Yu, Xiang, Zhou, and
  Bengio]{22}
Zhouhan Lin, Minwei Feng, Cicero~Nogueira dos Santos, Mo~Yu, Bing Xiang, Bowen
  Zhou, and Yoshua Bengio.
\newblock A structured self-attentive sentence embedding.
\newblock \emph{arXiv preprint arXiv:1703.03130}, 2017.

\bibitem[Raffel and Ellis(2015)]{20}
Colin Raffel and Daniel P.~W. Ellis.
\newblock Feed-forward networks with attention can solve some long-term memory
  problems.
\newblock \emph{arXiv preprint arXiv:1512.08756}, 2015.

\bibitem[S{\o}nderby et~al.(2015)S{\o}nderby, S{\o}nderby, Nielsen, and
  Winther]{7}
S{\o}ren~Kaae S{\o}nderby, Casper~Kaae S{\o}nderby, Henrik Nielsen, and Ole
  Winther.
\newblock Convolutional lstm networks for subcellular localization of proteins.
\newblock In \emph{International Conference on Algorithms for Computational
  Biology}, pages 68--80. Springer, 2015.

\bibitem[Vinyals et~al.(2015)Vinyals, Kaiser, Koo, Petrov, Sutskever, and
  Hinton]{9}
Oriol Vinyals, {\L}ukasz Kaiser, Terry Koo, Slav Petrov, Ilya Sutskever, and
  Geoffrey Hinton.
\newblock Grammar as a foreign language.
\newblock In \emph{Advances in Neural Information Processing Systems}, pages
  2773--2781, 2015.

\bibitem[Xu et~al.(2015)Xu, Ba, Kiros, Cho, Courville, Salakhutdinov, Zemel,
  and Bengio]{6}
Kelvin Xu, Jimmy Ba, Ryan Kiros, Kyunghyun Cho, Aaron~C Courville, Ruslan
  Salakhutdinov, Richard~S Zemel, and Yoshua Bengio.
\newblock Show, attend and tell: Neural image caption generation with visual
  attention.
\newblock In \emph{ICML}, volume~14, pages 77--81, 2015.

\bibitem[Yang et~al.(2016)Yang, Yang, Dyer, He, Smola, and Hovy]{21}
Zichao Yang, Diyi Yang, Chris Dyer, Xiaodong He, Alex Smola, and Eduard Hovy.
\newblock Hierarchical attention networks for document classification.
\newblock In \emph{Proceedings of the 2016 Conference of the North American
  Chapter of the Association for Computational Linguistics: Human Language
  Technologies}, pages 1480--1489, 2016.

\end{thebibliography}

\end{document}